\documentclass[sigplan,screen]{acmart}

\usepackage{microtype}
\usepackage{graphicx}
\usepackage{xspace}
\usepackage[caption=false]{subfig}

\usepackage{hyperref}
\usepackage{balance}

\usepackage{mathtools}
\usepackage{amsthm}
\usepackage{amssymb}

\usepackage{tabularx}

\usepackage[capitalize,noabbrev]{cleveref}

\usepackage{url}

\usepackage{color, colortbl}

\definecolor{brickred}{rgb}{0.8, 0.25, 0.33}
\colorlet{shadecolor}{gray!20}

\newcommand{\revision}[1]{#1}
\newcommand{\revisionfig}[1]{#1}

\newcommand{\niparagraph}[1]{\vspace{1pt}\noindent\textbf{#1}}

\newcommand{\neusight}{\textsc{NeuSight}}

\newcommand{\fig}[1]{Figure~\ref{#1}}

\newcommand{\micro}{Li et al.}

\newcommand{\cutlass}{CUTLASS}
\newcommand{\cudnn}{cuDNN}

\copyrightyear{2025}
\acmYear{2025}
\setcopyright{acmlicensed}
\acmConference[ASPLOS '25] {Proceedings of the 30th ACM International Conference on Architectural Support for Programming Languages and Operating Systems, Volume 1}{March 30--April 3, 2025}{Rotterdam, Netherlands.}
\acmBooktitle{Proceedings of the 30th ACM International Conference on Architectural Support for Programming Languages and Operating Systems, Volume 1 (ASPLOS '25), March 30--April 3, 2025, Rotterdam, Netherlands}
\acmISBN{979-8-4007-0698-1/25/03}
\acmDOI{10.1145/3669940.3707265}

\begin{document}

\title{Forecasting GPU Performance for Deep Learning Training and Inference}

\author{Seonho Lee}
\email{seonho.lee@gatech.edu}
\affiliation{%
\institution{Georgia Institute of Technology}
\city{Atlanta, GA}
\country{USA}
}

\author{Amar Phanishayee}
\authornote{Work done while at Microsoft Research.}
\email{aphanishayee@meta.com}
\affiliation{%
\institution{Meta}
\city{Seattle, WA}
\country{USA}
}

\author{Divya Mahajan}
\email{divya.mahajan@gatech.edu}
\affiliation{%
\institution{Georgia Institute of Technology}
\city{Atlanta, GA}
\country{USA}
}

\renewcommand{\shortauthors}{Seonho Lee, Amar Phanishayee, \& Divya Mahajan}

\begin{abstract}
%
Deep learning kernels exhibit a high level of predictable memory accesses and compute patterns, making GPU's architecture well-suited for their execution.
Moreover, software and runtime system for GPUs further enable optimizations that aim to better utilize the stream multiprocessors, on-chip bandwidth, multiple levels of cache hierarchy, and off-chip high-bandwidth memory.
In the context of deep learning, the entire space of models and GPUs is constantly evolving, as newer models emerge with simultaneous upgrades to the device.
However, access to newer GPUs is often limited, raising important questions about \emph{the performance of new model architectures on existing GPUs, existing models on new GPUs, and new model architectures on new GPUs.}
To address these questions, we introduce \neusight, a forecasting framework to predict the performance of a diverse range of deep learning models, for both training and inference, on unseen GPUs, without requiring actual execution of the target model on the target GPU.
The framework leverages both GPU hardware behavior and software library optimizations to estimate the end-to-end performance of these models.
We observe that prior work in this area suffers from high absolute error percentages when forecasting performance on unseen models and new GPUs, as they attempt to model the complex task of predicting the latency of a deep learning kernel on a GPU directly using a machine learning approach.
Instead, with \neusight, we decompose the prediction into smaller problems, while bounding the prediction through fundamental performance laws.
\neusight~decomposes a single deep learning kernel prediction into smaller working sets called tiles, which are executed independently on the GPU.
Tile-granularity predictions are determined using a machine learning approach and aggregated to estimate the end-to-end latency.
As such, \neusight~outperforms prior work across a variety of deep learning workloads and the most up-to-date GPUs.
It reduces the percentage error from 121.4\% and 30.8\% to 2.3\% in predicting the latency of GPT3 model for training and inference on H100, in comparison to state-of-the-art prior work, respectively, where GPT3 and H100 were not used to train any framework.

\end{abstract}

\sloppy



\maketitle 

\section{Introduction}

GPUs are the canonical platform to be deployed across any Artificial Intelligence (AI) cluster.
There are two main reasons for their widespread adoption for deep learning: (1) the parallel compute engines that mitigate the Von-Neumann overheads of general purpose CPUs and (2) closely knit software libraries that exploit the architecture for various Deep Neural Network (DNN) kernels~\cite{cudnn, cutlass}.
This trend has been further augmented by the availability of specialized AI servers, aiming to minimize the overheads of distributed execution~\cite{nvidia-dgx, amd-miserver}.

\begin{figure}[t]
\centering
\includegraphics[width=0.85\linewidth]{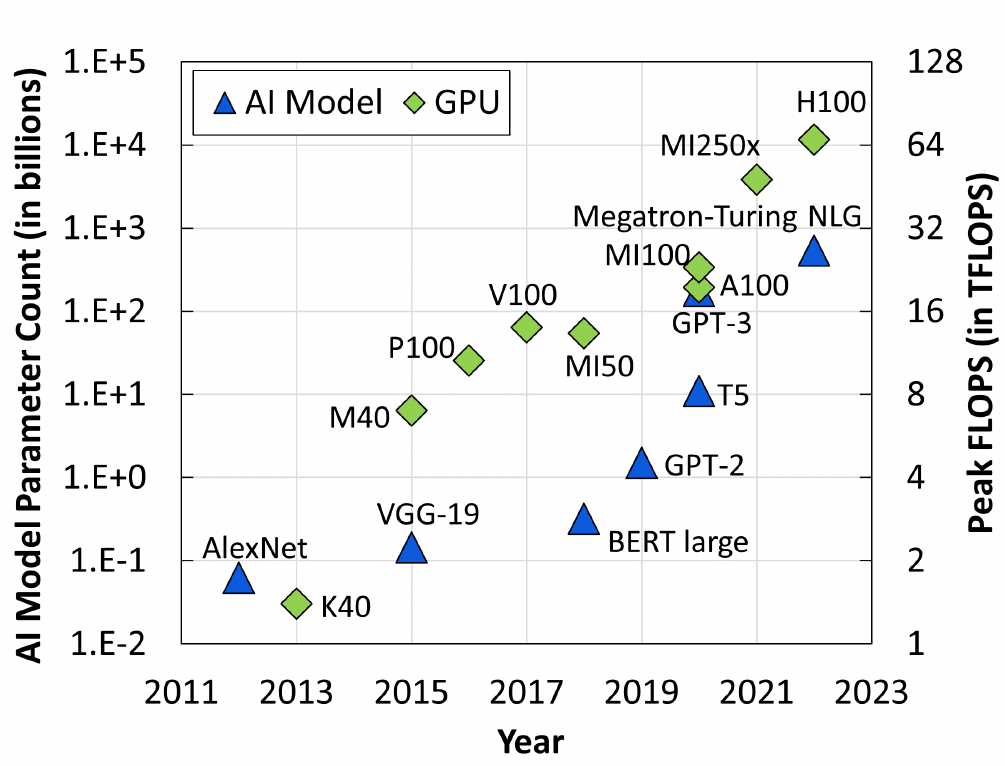}
\vspace{-2.5ex}
\caption{Growth of AI models and the compute and memory capacity of GPUs.~\cite{nvidia_gpus,amd_gpus,alexnet,vgg,bert,gpt2,t5,gpt3,megatronnlg} }
  \label{fig:model_gpu_growth}
  \vspace{-2ex}
\end{figure}

In this rapidly evolving field of AI, continuous iterations are introduced.
However, newer GPUs are limited in access - either due to the long lead time (of upto 52 weeks for H100~\cite{h100stretch}) or their prohibitively high cost. 
Currently, devising new models and the success of their execution is contingent on the availability of underlying hardware.
As shown in Figure~\ref{fig:model_gpu_growth}, since 2013, we have witnessed an exponential growth in model size, accompanied by a corresponding increase in the peak performance of GPUs.
In summary, the prevailing shift towards the widespread deployment of GPU-centric workloads raises a significant challenge: how can we closely estimate the performance of current and upcoming deep learning models on inaccessible GPUs?

The primary challenge in developing a solution capable of closely estimating GPU performance using only the abstract hardware architecture features lies in the fact that the performance of deep learning models on GPUs is intricately tied to low-level hardware details and software optimizations, such as those provided by DNN kernel libraries such as CuDNN and Cutlass.
Unlike hardware accelerators~\cite{tpuv4_isca, dnnweaver, tabla, dana, brainwave, eyerissv2}, GPUs exhibit complex interplay between micro-architectural interactions and execution characteristics, due L1 and L2 caches, DRAM, memory, I/O configurations, and warp scheduling.
Consequently, while there are various established works~\cite{sunstone, maestro} that can closely estimate the performance of DNN kernels on domain-specific accelerators, they are not suitable for GPU performance forecasting.
Nevertheless, previous works have aimed to leverage cycle-accurate simulators to provide detailed execution and performance estimates for GPUs~\cite{gpgpusim, accelsim, pka}.
However, these simulators need to be manually updated for each change in the GPU architecture. 
As such, developing a cycle accurate simulator is a time-consuming process and using them to simulate the simplest of the model can take a long time.
The most popular GPU simulator~\cite{accelsim, pka} can take up to 18 hours to simulate ResNet-50 with a batch size of 256~\cite{lietal}. ResNet-50 is relatively smaller compared to new-age complex networks such as the LLMs.
It will take a prohibitively long time to use this simulator to predict the performance of future deep learning models.

In this work, we devise a novel prediction mechanism based on machine learning, designed with the consideration of tile-based execution strategy on Stream Multiprocessors (SMs) in GPUs, and the compute and memory bounds of DNN kernels. 
%
%
We train our model using data collected from a set of readily available older generation GPUs and existing deep learning kernel information, which is then used to predict performance of newer models, both training and inference, on unseen GPUs.

Prior work in the domain of modeling GPU performance takes an analytical model, a linear regression-based model, or a neural network approach~\cite{habitat, lietal, calculon}.
We empirically observe that these solutions exhibit high percentage errors for newer models and GPUs, for both inference and training. 
They are limited for the proposed use case because they either: (1) use a rough analytical model without information about detailed GPU execution (hardware or software) (2) require execution on a specific GPU to extrapolate latency for newer GPUs~\cite{lietal, calculon}, or (3) are unable to capture the necessary architectural information of the GPUs~\cite{habitat}.
Additionally, prior work exhibits high percentage errors for newer models and unseen GPUs because they directly model the overall kernel latency using machine learning.
However, GPU execution, even of a single kernel, is a complex conflation of hardware execution across SMs and software optimizations, which often cannot be captured by a machine learning model trying to determine the latency of the kernel only using certain high-level features such as effective memory bandwidth, compute utilization, and peak flop.

Thus, we develop \neusight~which digresses from prior work as it leverages the insight that popular GPU libraries decompose deep learning kernels into multiple smaller working sets. These small working sets, or tiles, are then dispatched and executed on the GPU independently across the SMs.
\emph{Based on this observation we partition the end-to-end latency prediction into regular and more manageable sub-problems, predicting the latency at tile-granularity.}
However, note that the GPU execution is governed by a few performance laws - such as the performance cannot exceed the peak flops or peak memory bandwidth. 
\emph{We capture the above two aspects in our prediction and only use a machine learning model to determine the utilization of the device based on the DNN kernel and GPU properties.} 
The output of the MLP, i.e., utilization of the device, is subsequently used to predict the tile, operator, and the model level latency. 
Overall, the goal of this entire framework is to distill down to the properties of GPU execution, use machine learning to capture the runtime non-linear behavior, bound the results by fundamental performance laws, and gather the predictions for end-to-end latency predictions.
This approach ensures robustness in unseen scenarios, such as new GPUs and models.
To ensure prediction for large models that require multiple GPUs, we extend our approach to predict the latency of distributed execution within a server. To do so, we predict the latency of communication operators and integrate these with the per-device execution latencies determined by \neusight~to calculate the overall distributed performance.

The evaluation on a diverse set of GPUs (Nvidia H100, A100-80GB, V100, A100-40GB, P100, T4, L4) and deep learning workloads (BERT, GPT2, GPT3, OPT, Switch Transformer) shows that \neusight~demonstrates a percentage error of 8.9\%, compared to 140\% by MLP-based~\cite{habitat} and 60.8\% by linear regression-based~\cite{lietal} prior work, without being trained on H100, L4, and A100-80GB GPUs and trained only on a subset of GPT3 and OPT model dimensions.
When evaluated for distributed training on a 4-GPU server, H100 (DGX Box) and A100 (NVLink), \neusight~shows a modest percentage error of 5.4\% across GPT2 and GPT3 model.
The error presented throughout this paper is the mean absolute percentage error, calculated against actual measured latencies. For brevity, we henceforth refer to it as percentage error.

\section{Background}


\subsection{GPU Architecture}

%
The availability of parallelism in deep learning workloads and their suitability to GPU-architectures has led to co-optimization across various generations of workloads and GPUs.
%
%
While the terminology used in this paper is of NVIDIA GPUs, we showcase the applicability of \neusight~to both Nvidia and AMD GPUs. 
A GPU is comprised of a collection of identical building blocks known as Streaming Multiprocessors (SMs).
These SMs are connected to a L2 cache through an on-chip interconnection network. The L2 cache, in turn, is connected to memory through off-chip I/O. Each SM features multiple processing units and a small L1 cache. The regularity of the GPU architecture makes it scalable to tens of thousands of processing units, allowing it to effectively exploit parallelism in deep learning kernels.

\subsection{Deep Learning Execution}

\niparagraph{Kernel execution.}
Modern deep learning workloads consist of multiple layers, where each layer itself has multiple fine-grained kernels such as matrix multiplication or element-wise operators. 
In the context of this paper, DNN kernel is a tensor operator such as CONV, GEMM, Add, Softmax, that is executed entirely on the device once initiated.
Although the entire workload comprises thousands of kernels, due to the repetitive structure of the model, they can be decomposed into a few types of operators. 
\textit{These types are General Matrix Multiplication (GEMM), fully-connected layer, and point wise operators (ReLU, Tanh, Softmax, etc.). 
Hence, this work focuses on determining the execution of each kernel type on GPU to predict the latency as detailed in Section~\ref{sec:tiledexecution}.} 
%

\niparagraph{Per-device execution.}
Machine learning practitioners use frameworks such as PyTorch and TensorFlow~\cite{pytorch, tensorflow} to represent and deploy a model as a set of layers.
The dataflow of the kernels is managed by the framework and dispatched to the GPU.
The corresponding device library then executes each kernel and returns the results to the framework. 
For example, NVIDIA GPUs utilize the CuDNN library for kernel execution, abstracting the underlying hardware complexity from the user. 
\textit{In this paper, similar to prior work, we assume kernels execute sequentially on the device, which is also empirically validated for the mainstream machine learning frameworks \cite{daydream}.}
Thus, the predicted latency of each kernel is aggregated based on the dataflow of the model that executes on each device to determine the per-GPU execution latency.

\niparagraph{Distributed per-server execution.}
Due to the size of the model, training or inference is often executed in a distributed fashion.
Pipeline parallel and tensor model parallel are used to split a model across various devices, whereas data parallel is used to improve throughput~\cite{pipedream, gpipe, megatronnlg, dataparallel}.
Regardless, distributed execution warrants communication across devices such as send/receive for pipeline parallel and collective operations like all-reduce for tensor model parallel and data parallel. 
In this work, \neusight~supports prediction of latency when a model is distributed across GPUs within a single server. 
Within this server, GPUs can be connected through NVLink or like a DGX box to support the aforementioned parallelization techniques~\cite{nvlink, nvidia-dgx}.
The type of parallelism, its width, and the schedule is provided by the user as an input to \neusight.

\begin{figure}[t!]
\centering  
\subfloat[\label{fig:Habitat_motivational} Multi-Layer Perceptron Approach (Habitat)~\cite{habitat}]{\includegraphics[width=\columnwidth]{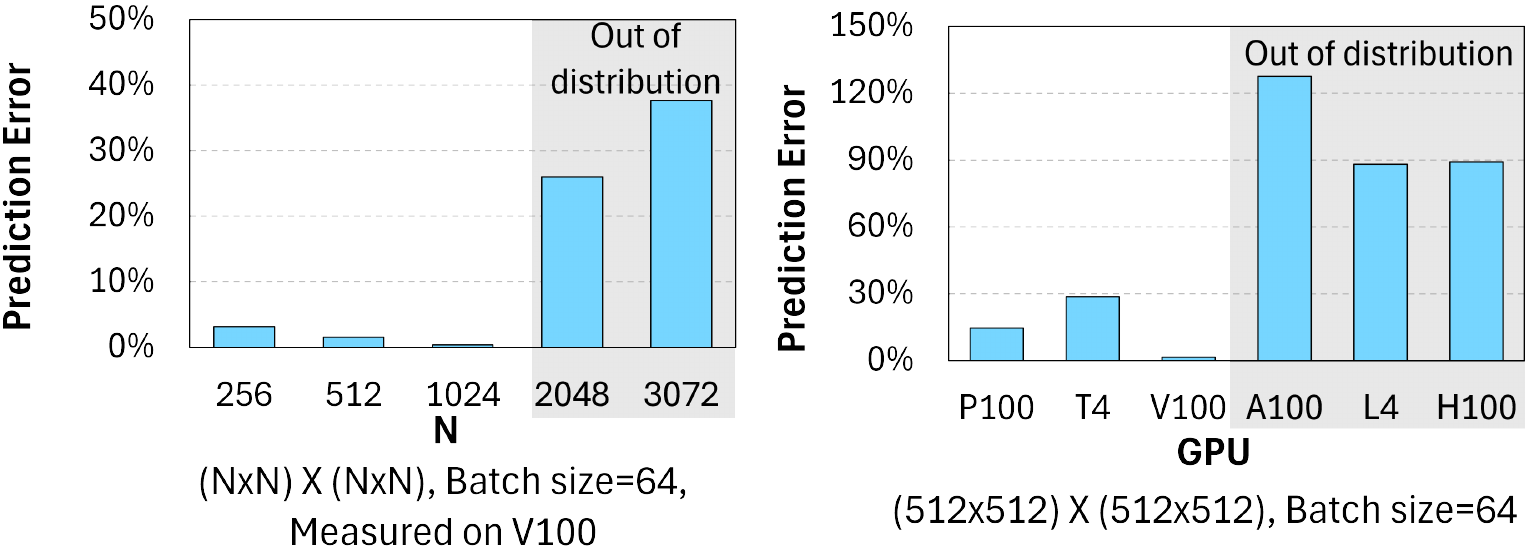}}
\\
\vspace{-1ex}
\subfloat[\label{fig:micro_motivational} Linear Regression-based Approach (Li et al.)~\cite{lietal}]{\includegraphics[width=\columnwidth]{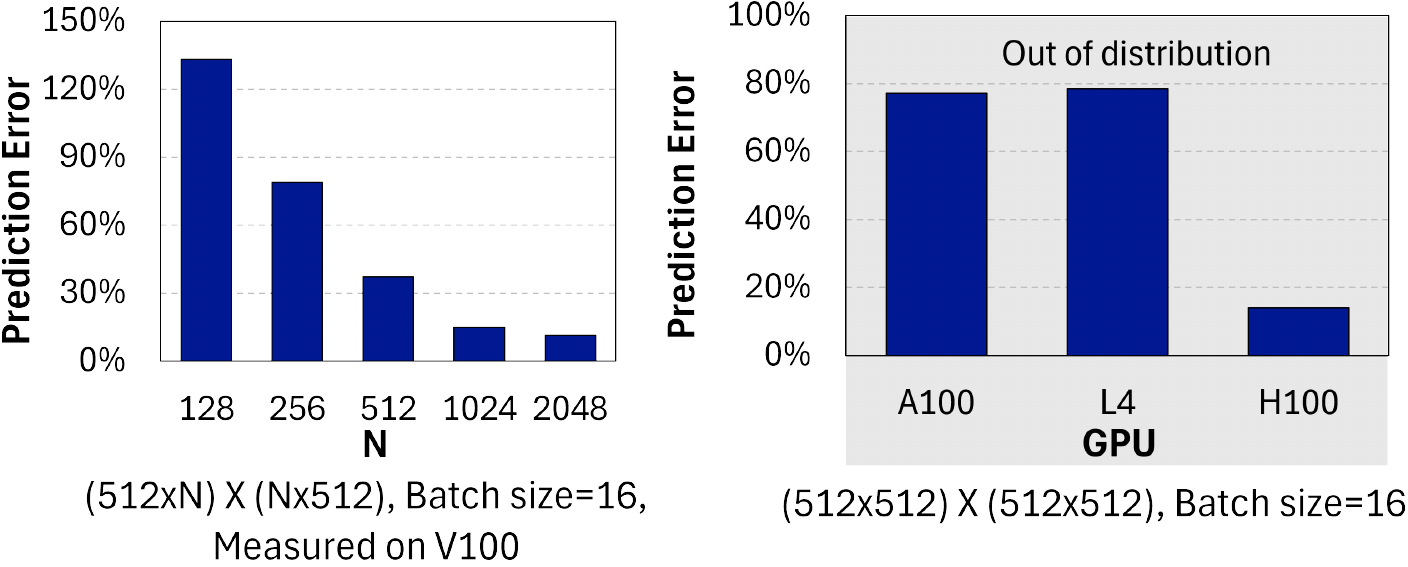}}
  \caption{Prediction error of prior work on BMM operator, reported in percentage error. Out of distribution dimensions and GPUs are highlighted. For these results, we trained Habitat and Li et al. models only up to V100, excluding any A100s, H100, and L4.}
\label{fig:motivational}
\end{figure}

\section{Motivation}

There is a growing need to predict GPU performance for both existing and upcoming models for a variety of reasons: (a) comparing performance across various devices, (b) utilizing estimates to identify GPUs that meet the performance requirements, and (c) evaluating the feasibility of deploying new model architectures on either existing or new GPUs.
We empirically validate prior work aimed at predicting the performance of GPUs by employing data-driven methods and observe that do not generalize well to new GPUs or new models.
These drive the insights for our framework, \neusight.

In this section, for motivation, we focus on a single kernel type Batched Matrix Multiplication (BMM), commonly found in DNNs. 
We present end-to-end comparisons (training and inference) in Section~\ref{sec:eval}.
We observe that even predicting performance of a single kernel on a GPU requires a careful consideration of hardware architecture and software optimization.
This insight is based on empirical validation of prior work employing machine learning to predict operator latency.

\subsection{Predicting Performance of Batched Matrix Multiplication Using prior work}
\label{motivation_eval}

We cover two most closely related prior works, Habitat~\cite{habitat} and a linear regression based approach~\cite{lietal} to estimate the latency of GPUs for deep learning.

Habitat classifies operators into two types: kernel-alike which use similar kernels across different GPU architectures, and kernel-varying which use different kernels across different GPU architectures. 
For kernel-alike operators (such as vector operators), it executes applications on an existing GPU to measure latency and scales the latency with the ratio of target GPU hardware configuration to that of existing GPU. 
As such, Habitat requires a GPU in-hand to estimate the latency of these operators.
For kernel-varying operators (such as BMM), they train multiple Multi-Layer Perceptron (MLP) models to predict the latency of different operators.
The MLP trains on the following GPU features: memory size, memory bandwidth, number of SM cores, Peak FLOPS, and kernel input and output dimensions as input, and the measured latency as the output.

\micro\cite{lietal}~proposes a linear regression model to predict performance based on FLOPs count of target kernel.
For each GPU, they collect multiple data points to construct a linear regression between latency and the number of FLOPs (determined based on the size of input/output matrices). 
They observe that achieved performance across different GPUs is usually linearly proportional to their memory bandwidth.
Thus, they construct a linear regression between the memory bandwidth of GPUs and achieved FLOPS performance, using this relationship to infer the performance of GPUs outside the training set.

\fig{fig:motivational} reports the prediction errors for these two works.
For Habitat, we utilize the vanilla MLP model provided by the authors. 
This model was trained on diverse GPUs predating 2018, including V100, 2080Ti, T4, RTX2070, P100, and P4000, with matrix dimensions of up to 1024 and a batch size less than 128. 
We aim to understand the applicability of this work on newer GPUs and larger dimensions common in newer models. 
As Figure~\ref{fig:Habitat_motivational} shows, Habitat faces challenges in generalizing to matrix multiplications with dimensions larger than 1024, resulting in a percentage error of up to 38\%.
Additionally, Habitat struggles with unseen GPUs, displaying a percentage error of up to 127\% on A100.

For \micro, we gather BMM data on P4, T4, P100, and V100.
We use linear regression to correlate the kernel's FLOP count with latency, following the paper's outlined procedures.
For GPUs outside the training set (A100, L4, and H100), we use this linear regression to predict the latency.
As Figure~\ref{fig:micro_motivational} shows, the linear regression assumption fails on matrix multiplications with smaller dimensions due to under-utilization of the GPU, when the latency does not scale linearly.
This work also faces challenges in generalizing to unseen GPUs because of its simplistic extrapolation method, i.e., using a linear ratio of memory bandwidth to infer the effective performance of unseen GPUs.

\begin{scriptsize}
\newcommand\ExtraSep
{\dimexpr\cmidrulewidth+\aboverulesep+\belowrulesep\relax}

\newcolumntype{?}{!{\vrule width 2pt}}
\setlength\extrarowheight{3pt}

\begin{table}[t]
\caption{Various ML models on predicting latency of BMM. Prediction errors are reported in percentage error.}
\vspace{-2ex}

\begin{tabular}{llll}
\hline
   \textbf{Predictor}                &  \textbf{Number} & \textbf{In-distribution}  & \textbf{Out-of-distribution}  \\
 \textbf{Architecture} & \textbf{of layers
} & \textbf{Prediction Error} (\%)      & \textbf{Prediction Error} (\%)          \\ \hline
\textbf{MLP}                & 8                & 28.0                       & 70.9                          \\ \cline{2-4} 
                   & 16               & 22.3                       & 81.4                             \\ \hline 
\textbf{Transformer}         & 3                & 22.3                       & 126.1                           \\ \cline{2-4} 
                   & 6                & 21.0                        & 86.4                          \\
\hline
\end{tabular}

\vspace{-2.5ex}

\label{tab:model_ablation}
\end{table}

\end{scriptsize}

\subsection{Predicting Performance of Batched Matrix Multiplication with Larger Predictors}
\label{sec:largermodels}

One potential solution to reduce the prediction error is to use larger models for the same problem, i.e., predicting the latency of an operator using a machine learning model with more layers and features.
We extend the study to use a larger and more complex model with an increased number of layers to capture a broader set of execution patterns.
To test the limitations of relying purely on large machine learning models, we add more layers to Habitat's MLP or alternatively use a transformer architecture from prior work, Prime \cite{prime}. 
We train these models using the same method specified in Section \ref{sec:ImplementationDetails}.
BMM operations with dimensions between 1 and 4096 are used as the test set.
Table \ref{tab:model_ablation} summarizes the prediction error averaged across in-distribution BMM operations (matrix dimensions less than 1024) and out-of-distribution BMM operations (some dimensions larger than 1024).
Despite improved performance with larger models, all the evaluated models still show high percentage error exceeding 70\% for out-of-distribution cases.

It is evident from the results that naively using machine learning techniques to predict kernel latency does not suffice, as they fail to capture the continuous performance improvements of GPUs resulting from both software and hardware optimizations.
This drives the key insight for \neusight: unlike prior work, it does not directly predict the latency of kernels using machine learning; instead, it distills the kernel execution into tiles, uses a machine learning model to predict the utilization of the device per tile, and then determines the latency based on performance bounds defined by GPU architecture. \neusight~only relies on architectural details readily available for any new or upcoming GPU.

\subsection{Other Related Works}

\niparagraph{Performance modeling using analytical frameworks.}
\revision{
Several works~\cite{hongetal, paleo, calculon, madmax} propose analytical models to predict GPU performance. 
Paleo~\cite{paleo}, Calculon~\cite{calculon}, and MAD-Max~\cite{madmax} decompose latency into computation and communication for distributed training. 
They estimate latency by measuring effective peak throughput with representative workloads on the target GPU, which requires access to the target GPU for estimation.
The primary goal of these works is to estimate performance for distributed execution and employ simpler analytical tools for per-device performance estimation. 
In contrast, \neusight~focuses on providing a higher-fidelity estimation tool for per-device performance prediction.
Nevertheless, \neusight~can be integrated with networking simulators to estimate for distributed settings.}

\niparagraph{Cycle accurate simulators.}
Cycle accurate simulators~\cite{gpgpusim} require detailed modeling of each GPU component.
For every new GPU, the simulator must be updated to support new architectural features, which might not be readily available at launch.
Proprietary architectural features can create a gap between real hardware and simulators. 
Additionally, cycle accurate simulators are not suitable for end-to-end ML application execution due to their long simulation times.
While certain works, such as sampled principal kernels~\cite{pka}, reduce simulation time, they still take at least hours to days to simulate modern DNN due to the need for modeling every detail of the hardware.

\niparagraph{Machine Learning based approaches.}
Due to the large number of compute lanes in GPUs, they often observe linear performance trends for compute intensive kernels, and hence many prior work leverage regression models to predict the performance of a GPU architecture~\cite{stargazer, wuetal, lietal}.
\revision{Some works require extracting low-level hardware features, such as hardware counter values, to make predictions.
However, these values are unavailable unless the device is accessible~\cite{halwpe, gatsim, hypbridgpuestimation}.}
Such approaches cannot capture the complex intricacies of hardware features and software optimizations such as tiling for deep learning and do not support cross-generational performance prediction with GPUs due to the lack of hardware.
Other works~\cite{driple} propose graph neural networks to predict resource consumption (GPU utilization, memory utilization, and network TX/RX throughput) in distributed deep learning.
However, this approach necessitates target GPUs for transfer training.

\niparagraph{CPU executions to predict speed-up on GPU.}
XAPP~\cite{xapp} executes the CPU code of a target algorithm to predict GPU speed-up. 
However, this approach assumes the GPU will execute the same algorithm but with higher parallelizability. 
In contrast, modern software optimizations for deep learning kernels are designed specifically for GPUs~\cite{flashattention, tvm}, diverging from CPU implementations.

\begin{figure}[t]
\centering
\subfloat[\label{fig:output_tiling} Tiling of output matrix.]{\includegraphics[width=0.35\textwidth]{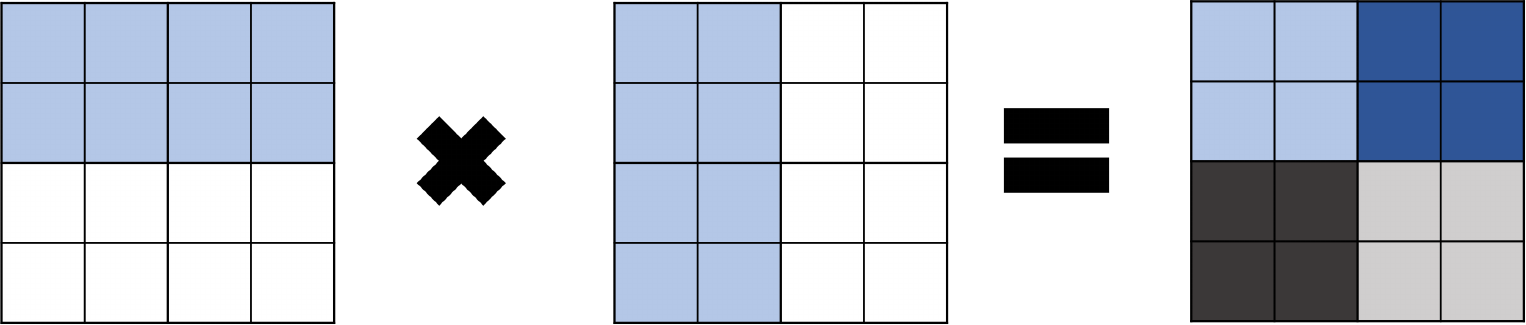}}
\\
\subfloat[\label{fig:smmapping}Mapping tiles to multiple SMs.]{\includegraphics[width=0.36\textwidth]{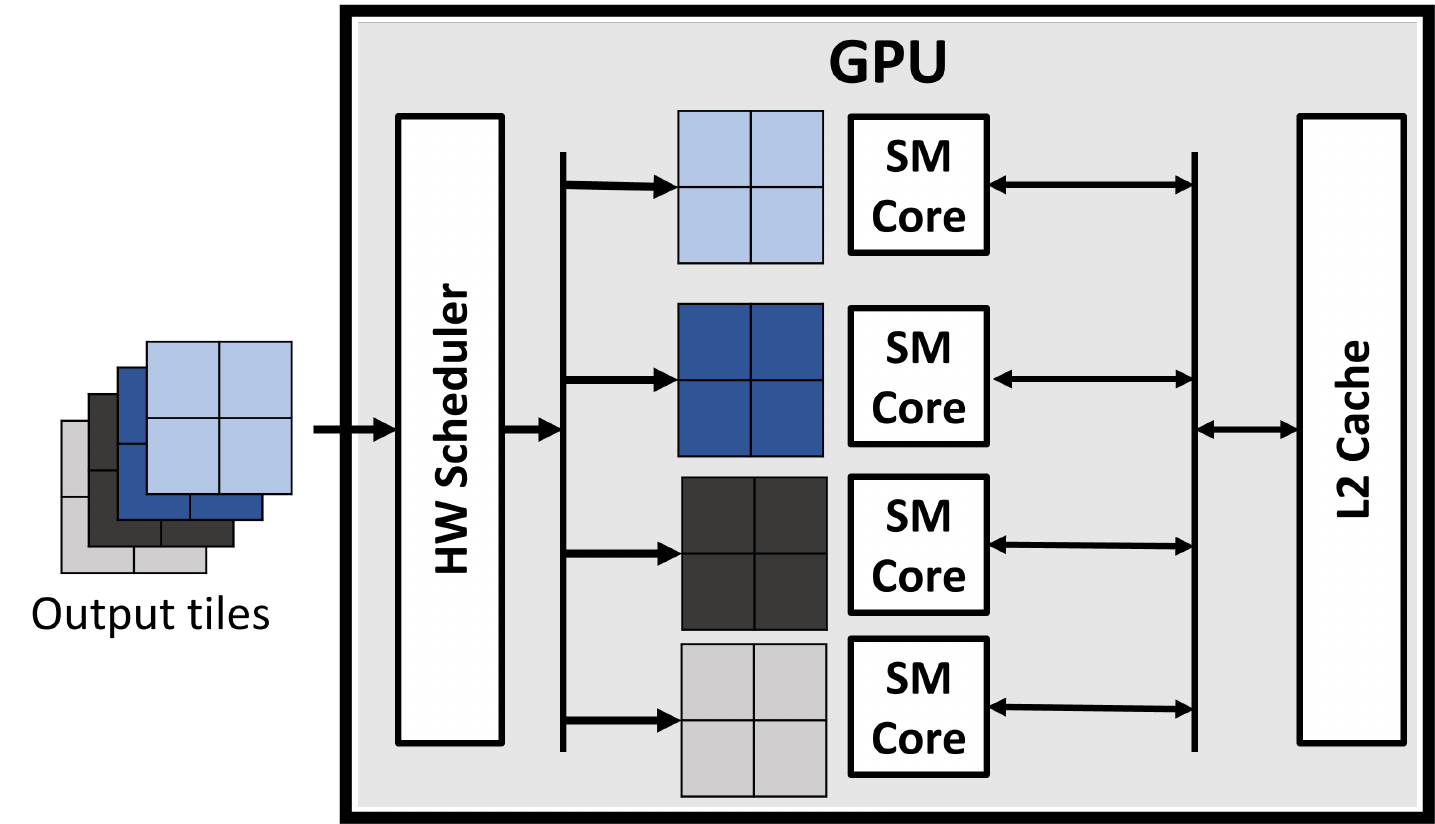}}
\vspace{-1.5ex}
\caption{Dataflow of a GEMM on a GPU. We assume multiplication between two 4x4 matrices and tile size of 2x2.}
\vspace{-1.5ex}
\label{fig:tiling}
\end{figure}

\section{\neusight~Forecasting}

\neusight~forecasts the latency of executing a model on multiple GPUs in a server in three steps. The first and main step forecasts the latency of a single kernel that executes on the GPU atomically.
The kernel leverages software and hardware optimizations for GPUs, namely tiling, parallel execution across SMs, on-chip caching, and HBM bandwidth. \neusight~uses architectural details such as memory size, memory bandwidth, number of SMs, L2 cache size, and peak FLOPS to predict the kernel's latency through a machine learning approach.
Our kernel estimation is inspired by how deep learning kernels are disassembled into multiple smaller tiles and mapped onto GPU hardware for execution. 
This approach differs from prior work, which relies solely on compute or memory utilization to predict latency.
Section~\ref{sec:tiledexecution} showcases the execution of a few representative DNN kernels on the GPU. \neusight~supports tensor operators such as General Matrix Multiplication (GEMM), element-wise operators such as addition and ReLU, and reduction-based operators such as Softmax and layer normalization.
%
In the second step, the per-operator latency is aggregated to predict the performance of the DNN workload based on its dataflow graph.
In the third step, detailed in Section~\ref{sec:distributed}, the performance of per-GPU execution is combined with predicted network latencies to determine the performance of a DNN executing across multiple GPUs in a server.

\subsection{Kernel Execution on GPUs}
\label{sec:tiledexecution}

\begin{figure}[t]
\centering
\includegraphics[width=0.3\textwidth]{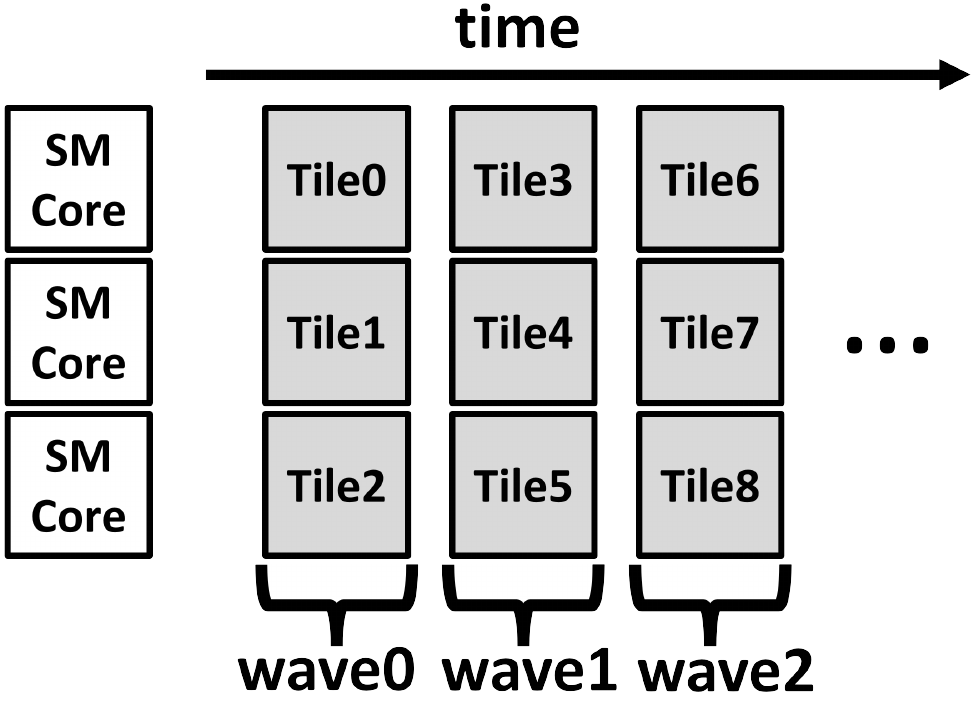}
\vspace{-1ex}
\caption{Each tile is distributed evenly across SMs and executed concurrently, in multiple number of waves. \cite{gpuperformancebackground}}
\vspace{-2ex}
\label{fig:wave}
\end{figure}

\niparagraph{Tiled execution.}
In this section, we focus on GEMM, a commonly found kernel in DNNs, but \neusight~can predict the latency of various operators to determine the end-to-end performance.
GEMM is a core building-block for many deep learning layers, such as fully-connected, convolution, recurrent neural network, or self-attention found in modern large language models. 
\fig{fig:tiling} shows a simplified view of a typical dataflow of GEMM on a GPU.
Modern GPU libraries execute GEMM by first partitioning the output matrix into multiple identical tiles (\fig{fig:output_tiling}). 
Each tile corresponds to a segment of the output matrix, loads the necessary input operands from input matrices, and computes the output elements. These tiles are mapped to each SM of the GPU and executed concurrently (\fig{fig:smmapping}). 
Typically the number of tiles which can be executed concurrently is limited by the number of SMs on the GPU, and the entire GPU kernel is executed in multiple waves of tile groups as depicted in \fig{fig:wave}.
Tiling strategy enables scalable execution of GEMM, by decomposing GEMM into multiple smaller workloads, which executes the identical computation on different input and output elements.

\niparagraph{Fundamental performance laws of GPUs.}
Runtime of each kernel is bounded by a few features of a GPU, such as peak memory bandwidth or peak FLOPs. 
Roofline analysis \cite{roofline} provides a straightforward method to estimate the approximate performance of a GPU kernel, taking into account the arithmetic intensity of the kernel. The intensity is computed by dividing number of flops ($flops_k$) by size of memory transactions ($mem_k$). The roofline bandwidth can be computed as follows:

\begin{equation}\label{eq:rooflinebw}
\footnotesize
    K = \frac{flops_k}{mem_k};~~ roofline_{BW} = min(K \times \revision{memBW_p}, flops_p)
\end{equation}

Where $flops_p$ is the peak FLOPs and \revision{$memBW_p$} is the peak memory bandwidth of GPU.
Roofline bandwidth represents the maximum achievable throughput of a kernel on a GPU.

\begin{figure}[t]
\centering
\includegraphics[width=0.4\textwidth]{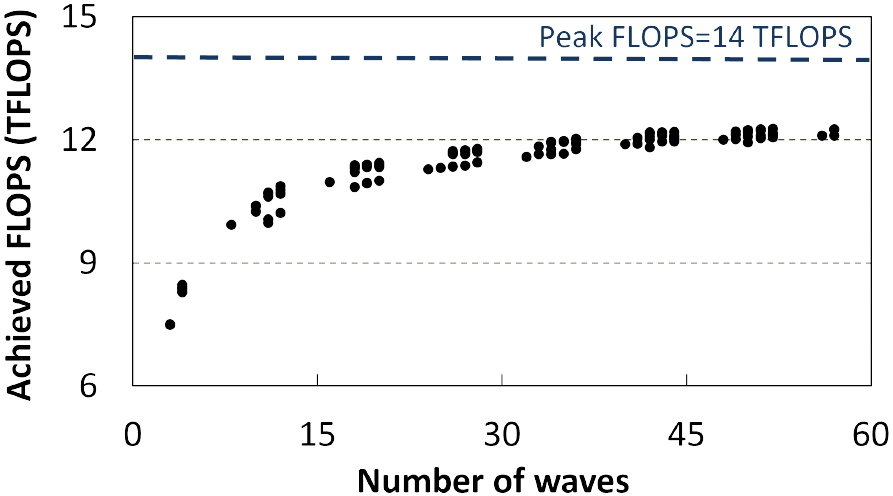}
\vspace{-1ex}
\caption{Performance of (256 × 256) × (256 × 256) matrix multiplication with varying waves on V100; sweeping the number of waves by changing the batch size from 1 to 300.}
\vspace{-1.5ex}
\label{fig:latencyhiding}
\end{figure}

\niparagraph{Latency hiding in GPU.}
\revision{
Although the roofline bandwidth represents the theoretical maximum bandwidth, each kernel does not always fully utilize the available compute or memory bandwidth.
Table \ref{tab:gpuutil} shows the hardware utilization of H100 when executing matrix multiplication, with a sequence length of 512 and a hidden dimension of 64, seen in BERT. 
The table shows that it is still common in machine learning models to not fully utilize compute or memory bandwidth.}

\revision{Another performance bound in GPU execution is the use of numerous threads to harness parallelism, achieving significantly higher throughput compared to traditional CPU architectures.
GPUs can achieve the aforementioned peak hardware performance ($roofline_{BW}$) only when there are enough available software threads to effectively mitigate latency arising from data dependency or memory stalls by scheduling independent threads. 
Consequently, GPU performance relies heavily on the level of parallelism in the kernels, which is directly proportional to the number of threads that can be independently scheduled in each SM.
\fig{fig:latencyhiding} illustrates the correlation between the number of waves available in each SM and the achievable throughput. 
As the number of waves increases, each SM is assigned more waves, resulting in a greater number of threads per SM. This allows for effective hiding of stall latency in each thread, thus leading to higher performance.}

\begin{scriptsize}
\newcommand\ExtraSep
{\dimexpr\cmidrulewidth+\aboverulesep+\belowrulesep\relax}
\newcolumntype{?}{!{\vrule width 2pt}}
\setlength\extrarowheight{3pt}

\begin{table}[t]

\centering
\caption{Measured compute utilization of H100 GPU when executing (512x64) x (64x512) matrix multiplication across various batch sizes.}
\revisionfig{
\begin{tabular}{lllllll}
\hline
\textbf{Batch Size}                                 & 32     & 64     & 128    & 256    & 512\\ \hline
\multicolumn{1}{c}{\textbf{Peak FLOPS Utilization}} & 53.2\% & 70.7\% & 69.4\% & 72.3\% & 86.0\%\\ \hline
\end{tabular}
}
\label{tab:gpuutil}
\end{table}

\end{scriptsize}

\subsection{Kernel-wise Prediction}
%
%
In predicting the latency of each kernel, we leverage the execution insights discussed above: tiled execution and fundamental GPU performance laws.

\niparagraph{Tile-granularity prediction.}
For each DNN kernel we break it down into multiple identical tiles, instead of directly predicting the latency of the kernel. 
This insight is motivated from prior work~\cite{habitat}, where a machine learning model is unable to capture the necessary hardware level and software optimizations on out of distribution dimensions and GPUs. 
This implies that there are more intricate execution patterns although consistent across the GPUs, but are not easily captured by machine learning models.
Thus, we use the machine learning model to make a prediction per tile, and assemble these predictions to determine the latency of the whole operator. 
This decomposition simplifies the problem into more manageable sub-problems, allowing the machine learning model to be trained on smaller and more straightforward tasks. 
For example, the typical tile dimensions used by GEMM library ranges from 32 to 256, which is still much smaller than the dimensions of GEMMs in modern DNN workloads.
To compute the latency of the entire kernel from the predictions for each tile, we employ the following: 

\footnotesize
\begin{equation}\label{eq:num_tiles}
    num_{tiles} = {\Pi^{N}_{i=1}}\lceil\frac{x_i}{t_i}\rceil
\end{equation}
\begin{equation}\label{eq:num_waves}
    num_{waves} = \lceil\frac{num_{tiles}}{num_{sm}}\rceil
\end{equation}
\begin{equation}\label{eq:perop_latency}
    PerOpLatency = PerTileLatency \times num_{waves}
\end{equation}
\normalsize 

Variables $x_i$ and $t_i$ are the dimensions of the output and the tile, respectively, $N$ is the number of dimensions of the output, and $num_{sm}$ is the number of SMs available in GPU. 
The tile dimensions are determined by metadata obtained with PyTorch Profiler discussed in Section~\ref{sec:ImplementationDetails}. 
\revision{In Equation \ref{eq:perop_latency}, we assume that each SM executes one tile at a time. 
The latency of the kernel typically scales linearly with the number of waves of tiles, and can be modeled as sequential execution of waves of tiles \cite{nvidiawave}.
As GPUs can execute threads from multiple tiles concurrently, this overlap is accounted for and abstracted in Equation \ref{eq:bwutil}, which will be discussed in the following subsection.}
In the next subsection, we discuss how we can obtain the $PerTileLatency$.

\niparagraph{Imposing performance laws to tile prediction.}
$PerTileLatency$ relies on the compute utilization of the SM and the achieved bandwidth of the kernel ($achieved_{BW}$), and is physically bounded by $roofline_{bw}$. This relationship is captured by the following equations:

\footnotesize
\begin{equation}\label{eq:pertile_latency}
PerTileLatency = \frac{flops_{tile}}{achieved_{BW}}
\end{equation}
\begin{equation}\label{eq:abw}
achieved_{BW} = roofline_{BW} \times utilization
\end{equation}
\normalsize

Next we need to predict GPU's utilization.
We use the observation that a GPU approaches the maximum achievable throughput as the number of available threads in the workload increases (as shown in \fig{fig:latencyhiding}). 
%
%
We formulate the correlation between number of waves and utilization as a machine learning model.
This is because, an ML model can capture the non-linear behavior between kernel latency and GPU characteristics and kernel properties such as operator type, tile dimensions, and arithmetic intensity.
We task MLPs with predicting the coefficients of the following equation:

\footnotesize
\begin{equation}\label{eq:bwutil}
utilization = alpha - \frac{beta}{num_{waves}}
\end{equation}
\begin{equation}\label{eq:alpha_beta}
alpha, beta = \sigma(MLP (input\_features))
\end{equation}
\normalsize

This equation models utilization, which increases as the number of waves increases due to the negative beta factor but is capped by alpha.
We apply a sigmoid function to the outputs of MLPs to bound the predicted utilization value to below 1.
This approach constrains the predictions to adhere to fundamental performance laws, enhancing its overall predictive capabilities even when faced with unseen inputs.

\subsection{Machine Learning Model to Predict Utilization}
We use an MLP with multiple fully-connected layers as our predictor. 
We train five distinct MLPs, each tailored for batched matrix multiplication, fully-connected layers, element-wise operators, softmax, and layer normalization. 

\niparagraph{Model architecture.}
Each MLP comprises 8 hidden layers, each with 512 hidden units, similar to prior work \cite{habitat}. First-layer converts the input feature vectors into 512-dimensional hidden vector, and the last layer converts the 512-dimensional hidden vector to single-dimensional output. ReLU is used as the activation function and applied at the end of every layer. This simple and small model architecture allows for fast but still accurate predictions, especially when combined with the GPU execution flow and performance scaling laws in the \neusight~framework.

We use memory size and bandwidth, peak FLOPS, and L2 cache size as GPU features for this modeling. 
We use these features because: (1) they are publicly available even for newer GPUs, (2) bandwidth and peak FLOPs determine the performance bounds of I/O bound and compute bound operators, and (3) on-chip memory size (L2 cache) and off-chip memory size (such as High Bandwidth Memory or DRAM) determine which underlying implementation will be used by the libraries such as \cudnn~and \cutlass~to execute the DNN operators.
For instance, information for the H100 was made available through NVIDIA's public documentation after its launch \cite{h100_whitepaper}.
For the Blackwell architecture, the latest NVIDIA GPU architecture announced~\cite{blackwell}, details on memory size, bandwidth, and peak FLOPs are already available. Information on the number of SMs and L2 cache size is unavailable, but based on prior years, should become available closer to the release date or shortly after the GPU is released.

Instead of directly feeding these input features to the MLP, we pre-process them through the following steps.
First, as we are making predictions on a tile granularity, with each tile mapped to a single SM, we divide the device resources by the number of SMs to determine per-SM resources (e.g., per-SM peak FLOPS).
Second, since the MLP predicts achievable throughput, we feed it with the utilization of each hardware resource. 
The five features shown in Table \ref{tab:features} are fed to the MLP as an input.

\begin{scriptsize}

\newcolumntype{?}{!{\vrule width 2pt}}
\setlength\extrarowheight{5pt}

\begin{table}[h!]
\centering
\caption{List of input features for predicting the utilization.}
\resizebox{0.8\columnwidth}{!}
{
\begin{tabular}{ c | c }
Input Features & Unit \\
 \hline
$\frac{FLOPs Per Tile }{Peak FLOPS Per SM} $   & $\frac{GFLOPS}{TFLOPS/s}$ \\
$\frac{Memory Per Tile}{Memory BW Per SM}$     & $\frac{MB}{GB/s}$     \\
$\frac{num_{waves}~\times~Memory Per Tile}{L2 Cache Size Per SM}$ & $\frac{MB}{KB}$       \\
$\frac{num_{waves}~\times~Memory Per Tile}{Memory Size Per SM}$   & $\frac{MB}{MB}$       \\
$\frac{FLOPs Per Tile~/~Memory Per Tile}{Peak FLOPS~/~Memory BW}$     & $\frac{GFLOPS~/~MB}{(TFLOPS/s)~/~(GB/s)}$ 

\end{tabular}
}

\label{tab:features}

\end{table}
\end{scriptsize}

\begin{figure*}
    \centering
    \includegraphics[width=0.95\textwidth]{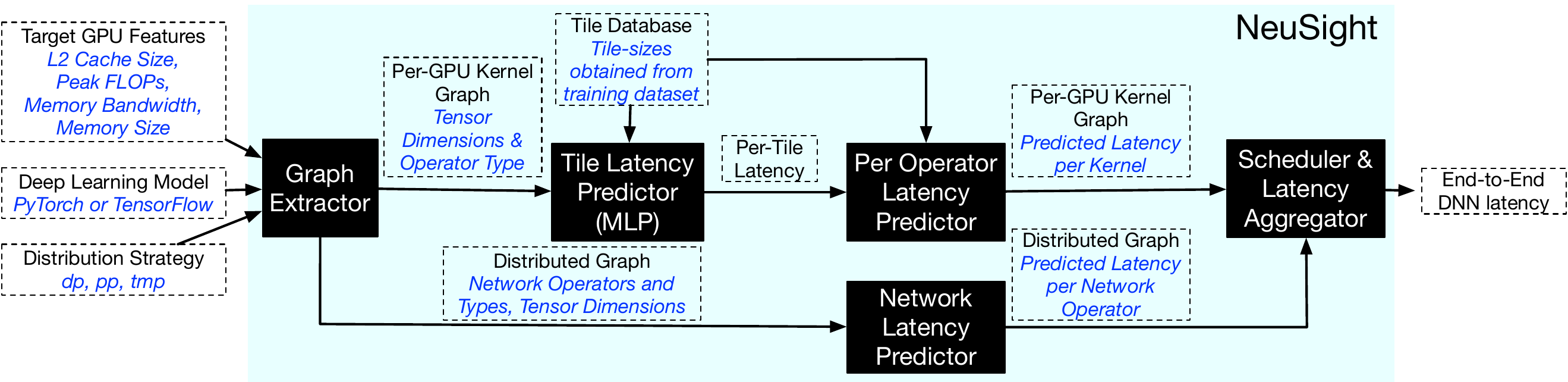}
    \caption{Overall workflow of \neusight. The Tile Latency Predictor uses an MLP to determine the utilization and predicts the kernel latency. This prediction is aggregated for per-device and distributed execution latency forecasting.}
    \label{fig:workflow}
    \vspace{-1ex}
\end{figure*}

\niparagraph{Predictors for various types of kernels.}
\neusight~includes multiple specialized predictors to support various types of DNN kernels. 
There are five MLPs to predict the utilization for BMM, fully-connected layers, element-wise operators, softmax, and layer normalization. 
For unseen operators, we consider them memory-bound and estimate latency by dividing the memory requirement by memory bandwidth. These diverse predictors enable \neusight~to handle various model architectures.

\subsection{Support for Operator Fusion}

Modern ML frameworks \cite{tvm, pytorch} accelerate the execution of memory-bound kernels by fusing multiple consecutive operators and eliminating the data movement from external memory.
\neusight~supports the prediction of the fused operators. 
When multiple consecutive vector kernels are fused, \neusight~accumulates the  FLOPs count of each operator, but discards the memory size required for the intermediate results between kernels. 
\neusight~uses the updated FLOPs count, memory size of the fused operator, and the metadata including tile size of the first operator to be fused to make predictions with its MLP.
For instance, in the GPT-2 architecture, the element-wise addition used for residual connections in each layer can be fused with subsequent layer normalization. \neusight~can predict the latency of this fused operator accumulating the FLOPs count of both operators and discarding the memory size required for the output of the element-wise addition and the input of the layer normalization, and using the predictor for element-wise addition.
Another common form of operator fusion is combining the GEMM operator with  activation functions. \neusight~supports this fusion using the BMM or fully-connected layer predictor and adjusting the input compute count and memory size.

\section{\neusight~Workflow}

\neusight~forecasts the end-to-end latency of a deep learning model executing on a single GPU or a multi-GPU server in three steps: (1) forecasting the performance of per-kernel execution on the GPU as detailed above, (2) combining these kernel-level estimates based on the dataflow graph of the DNN to determine the per-GPU latency, and (3) estimating collectives and network operations and integrating them with the per-device execution latency to determine the performance on a GPU server executing distributed training or inference.
Both training and inference graphs are supported.

\fig{fig:workflow} shows the overall workflow of \neusight.
It takes the description of the target deep learning model in PyTorch and extracts the operator/kernel graph using the Torch.fx library \cite{torchfx}. 
%
%
It records the metadata of each kernel, including operator type and input/output tensor dimensions. Based on the metadata and tile size, each operator is annotated with predictions made by our kernel-level predictor. The per-GPU latency is computed by accumulating the predicted latency of each kernel executing on a single device, which aligns with device execution where kernels are executed sequentially on the GPU \cite{daydream}.

\neusight~also takes as input the distribution strategy, data parallel width, pipeline parallel depth and schedule, or tensor model parallel width. \emph{It is important to highlight that the focus of \neusight~is to effectively predict the performance of DNNs on new or unseen GPUs.} Predicting the performance of each kernel is the complex and unsolved aspect tackled by~\neusight, as it involves addressing the non-linearity of GPU software optimizations and hardware execution.
%
For completeness, we provide a solution that can forecast the latency of distributed execution on a multi-GPU server. 
The goal of this forecasting is to enable researchers to determine latency of a relatively larger model that often cannot fit on a single device due to memory constraints.
This allows \neusight~to provide accurate and reliable predictions for the performance of the compute/memory intensity of the DNN while avoiding the variability introduced by complex multi-server network operations.
"Prior work~\cite{topoopt, astrasim} already models complex network topologies, which researchers can leverage alongside \neusight."

\subsection{Forecasting for Distributed Execution}
\label{sec:distributed} 

\neusight~supports a single multi-GPU server execution similar to the ones connected through NVLink or DGX Box configurations. 
For each form of parallelism, \neusight~adds the corresponding network operators to the DNN graph. Pipeline parallelism involves sending and receiving activations between GPUs in subsequent pipeline stages, while data parallelism and tensor model parallelism perform allreduce operations on gradients and activations, respectively.

\niparagraph{Estimating performance of network operators.}
\neusight~estimates the latency of ring-based allreduce and peer-to-peer send/receive based on the bandwidth of the network link. 
We measure the link bandwidth of an existing system and calculate its network link utilization. Using this data, along with the peak link bandwidth of the target system, we estimate the latency of the collective operators on the target system.
The estimated latency of network operators is combined with the predicted latency of per-GPU DNN graph to predict the end-to-end latency of distributed  execution. \neusight~uses the following methods for each parallelization strategy. 

\niparagraph{Pipeline parallel.} 
Pipeline parallel requires forward and backward pass micro-batches to be interleaved to achieve high utilization as per the schedules proposed by prior work~\cite{gpipe, pipedream, pipedream_flush}.
For \neusight, the user provides the pipeline parallel schedule and the framework stitches the execution across various devices to determine the overall latency of the model executing on a server.
Currently, \neusight~inserts the necessary pipeline bubbles between the forward and backward pass based on GPipe \cite{gpipe} schedule, however can be easily extended to other schedules. 
\neusight~estimates the latency of these pipeline bubbles based on the size of the microbatch, the number of GPUs, and the latency of send-receive operations. 

\niparagraph{Tensor model- and Data- parallel.}
\neusight~inserts all-reduce operators into the DNN graph to synchronize the activations (in tensor model parallel) or gradients (in data parallel) across GPUs. 
The framework currently supports Megatron tensor model strategy~\cite{megatronnlg}.
The latency of ring-based all-reduce is estimated and aggregated with the compute latency of kernels to determine the end-to-end latency.

\section{Evaluation}
\label{sec:eval}

\subsection{Implementation Details} 
\label{sec:ImplementationDetails}

\begin{footnotesize}
\newcommand\ExtraSep
{\dimexpr\cmidrulewidth+\aboverulesep+\belowrulesep\relax}

\newcolumntype{?}{!{\vrule width 2pt}}
\setlength\extrarowheight{3pt}

\begin{table}
\centering
\caption{List of GPUs used to train and test the frameworks. \revision{GPUs reserved for the test set are highlighted in grey.}}
\resizebox{1.0\columnwidth}{!}
{
\begin{tabular}{ c | c | c | c | c | c | c | c }

 \hline
\textbf{Vendor} & \textbf{GPU} & \textbf{Year} & \textbf{Peak FLOPS} & \textbf{Memory} & \textbf{Memory} &  \textbf{\#} & \textbf{L2 Cache}  \\
\textbf{} & \textbf{} & \textbf{} & \textbf{(TFLOPS)} & \textbf{Size (GB)} & \textbf{BW (GB/s)} &  \textbf{SMs} & \textbf{(MB)}  \\
 \hline
NVIDIA 
    & P4    & 2016 & 5.4  & 8 & 192 & 40  & 2   \\
    & P100  & 2016 & 9.5  & 16& 732 & 56  & 4   \\
    & V100  & 2017 & 8.1  & 32& 900 & 80  & 6   \\
    & T4    & 2018 & 14.1 & 16& 320 & 40  & 4   \\
    & A100-40GB  & 2020 & 19.5 & 40 & 1555 & 108 & 40  \\
\rowcolor{shadecolor}
& A100-80GB  & 2020 & 19.5 & 80 & 1935 & 108 & 40  \\
\rowcolor{shadecolor}
& L4     & 2023 & 31.3 & 24 & 300 & 60  & 48 \\
\rowcolor{shadecolor}
& H100   & 2022 & 66.9 & 80 & 3430& 132 & 50 \\
\hline 
\hline
\revision{AMD} & \revision{MI100}    & \revision{2020} & \revision{23.1,  46.1(Matrix)} & \revision{32} & \revision{1230}  & \revision{120} & \revision{8}   \\
    & \revision{MI210}  & \revision{2021} & \revision{22.6, 45.3(Matrix)} & \revision{64} & \revision{1640}  & \revision{104} & \revision{16}   \\
    \rowcolor{shadecolor}
    & \revision{MI250 (per die)}  & \revision{2021} & \revision{22.6, 45.3(Matrix)} & \revision{64} & \revision{1640}  & \revision{104} & \revision{16}   \\
\hline

\end{tabular}
}

\label{tab:gpus}

\end{table}
\end{footnotesize}

\begin{footnotesize}
\newcommand\ExtraSep
{\dimexpr\cmidrulewidth+\aboverulesep+\belowrulesep\relax}
\newcolumntype{?}{!{\vrule width 2pt}}
\setlength\extrarowheight{3pt}
\begin{table}
\centering
\caption{Workloads evaluated. Table details model complexity through parameter size and architecture configurations.}

\resizebox{1.0\columnwidth}{!}
{
\vspace{-1.5ex}
\begin{tabular}{ l | c | c | c | c | c | c }

 \hline
\textbf{Model} &\textbf{Year}  & \textbf{Parameter} & \textbf{\# of} & \textbf{\# Attention} & \textbf{Hidden} & \textbf{Sequence}\\
\textbf{} &\textbf{}  & \textbf{Size} & \textbf{Layers} & \textbf{ Heads} & \textbf{Dimensions} & \textbf{Length}\\
 \hline
 BERT Large & 2018 & 340M  & 12 & 16 & 760  & 512  \\
GPT2 Large  & 2019 & 774M  & 36 & 20 & 1280 & 1024 \\
GPT3 XL     & 2020 & 1.3B  & 24 & 24 & 3072 & 2048 \\
OPT 1.3B    & 2022 & 1.3B  & 24 & 24 & 2048 & 2048 \\
GPT3 2.7B   & 2020 & 2.7B  & 32 & 32 & 2560 & 2048 \\
SwitchTrans  & 2021 & 5.3B  & 24 & 32 & 1024 & 512 \\
\hline

\end{tabular}

\label{tab:workloads_apdx}

}

\end{table}
\end{footnotesize}

\niparagraph{Hardware.}
\revision{
Table \ref{tab:gpus} shows the GPUs used to evaluate \neusight. We use 5 Nvidia GPUs and 2 AMD GPUs released from 2016 to 2021 for training our MLP predictor. 
AMD GPUs have a dedicated datapath for FP32 matrix multiplication, and peak FLOPS for matrix multiplication as shown\cite{cdna2}.
Additionally, we use 2 Nvidia GPUs released in 2022 and 2023 as out-of-distribution GPUs.}
%
%
For all the final results, A100-40GB variant was used for training but the A100-80GB variant with larger memory and higher memory bandwidth was used as the testset.

\niparagraph{Generating the training dataset.}
We use the 5 training set GPUs shown in Table~\ref{tab:gpus} to collect data to train our MLPs.
Latency is measured by running each operator 25 times and averaging the results.
Input tensors are initialized using a normal distribution.
\revision{
All measurements were conducted in fp32 format, using PyTorch 2.1.0 compiled with CUDA Library 12.1 for NVIDIA GPUs, and PyTorch 2.4.1 compiled with ROCm 6.1 for AMD GPUs.
}
The collected data covers the following operators:

\begin{itemize}
    \item \textbf{Batched Matrix Multiplication - 87,627 data points} Batch size and dimensions from 1 to 1024.
    
    \item \textbf{Fully-connected Layer - 32,256 data points.} Batch size from 1 to 8192, input and output sizes from 1 to 65,536.
    
    \item \textbf{Element-wise operators - 26,066 data points.} Batch size from 512 to 16,384, vector size from 512 to 4096 for addition, division, multiplication, GELU, ReLU, and Tanh.
    
    \item \textbf{Softmax - 1,807 data points.} Batch size from 4096 to 16,384, vector size from 512 to 4096.
    
    \item \textbf{Layer normalization - 1,501 data points.} Batch size from 4096 to 16,384, vector size from 512 to 4096. 
\end{itemize}

We reserve 20\% of this sample dataset for validation. For other operators, such as the embedding layer, which do not contribute significantly to the end-to-end latency (e.g. less than 0.1\% for BERT Large), we assume that they are memory-bound operators and estimate latency by dividing the working memory size by the memory bandwidth of the GPU.

\begin{figure*}[t!]
\centering
\vspace{-2ex}
\subfloat[\label{fig:eval-inf} Inference latency prediction error]{\includegraphics[width=1\textwidth]{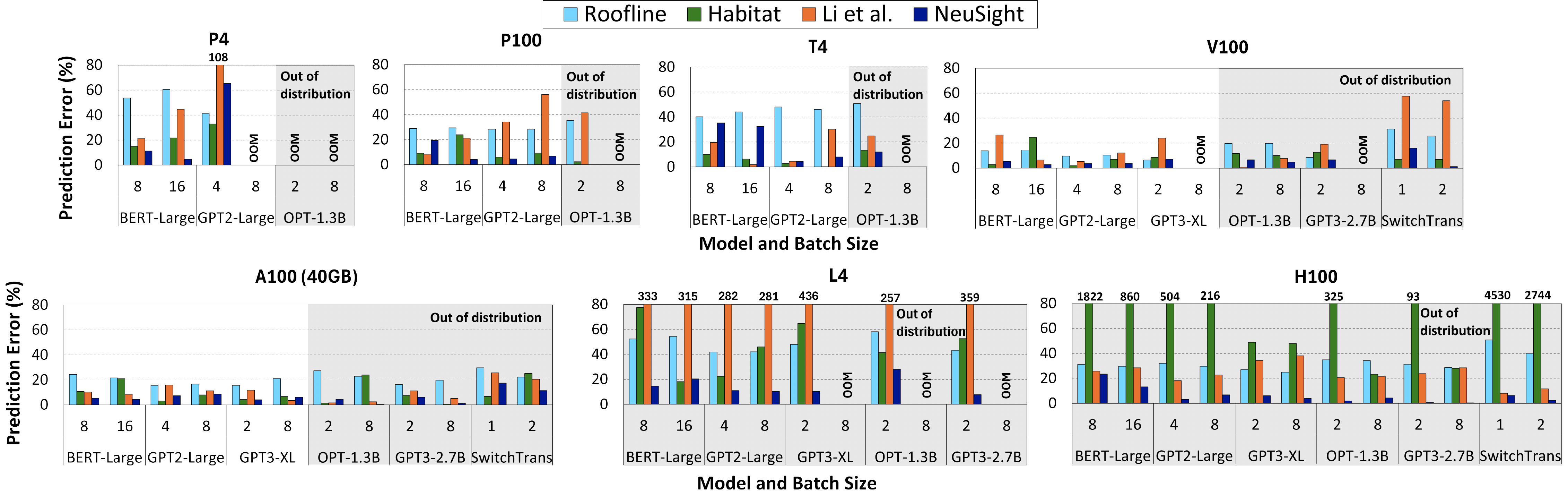}}
\\ \vspace{-1ex}
\subfloat[\label{fig:eval-train}Training latency prediction error]{\includegraphics[width=1\textwidth]{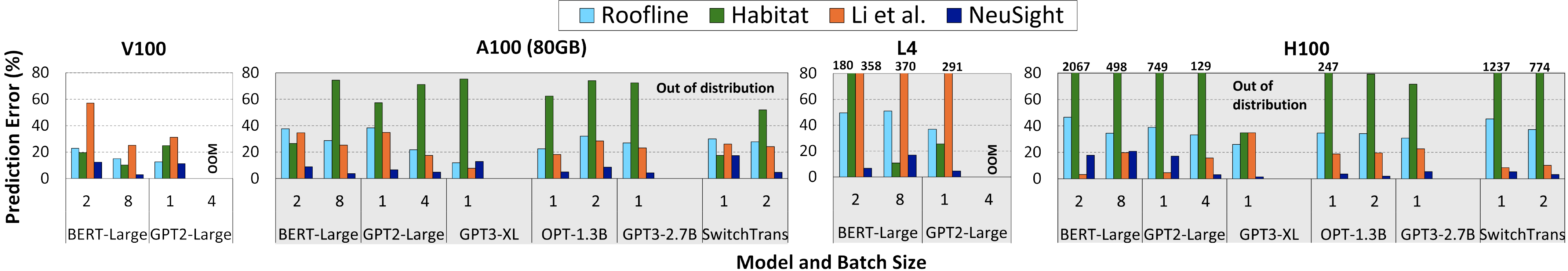}}
\vspace{-2ex}
\caption{The inference and training latency prediction percentage error of \neusight~and baselines with different batch sizes. Out of distribution GPUs and models are highlighted. Certain models resulting in OOM are omitted due to space constraints.}
\vspace{-1.5ex}
\label{fig:main_eval}
\end{figure*}

\niparagraph{Baselines.} We compare \neusight~with three baselines: (1) the roofline analysis, most widely used analytical performance estimation method, (2) Habitat~\cite{habitat}, a recent work that uses an MLP to predict GPU performance, and (3) \micro~~\cite{lietal}, the linear regression based method to predict the latency of deep learning models.
For fair comparisons, we train the baselines with the up-to-date training dataset as \neusight.
For Habitat, we train the MLP following the methodology presented in their paper, but with the larger MLP for higher accuracy (Section~\ref{sec:largermodels}).  
For element-wise operators, Habitat requires a reference latency of the  operator on an existing GPU.
We measure the latency of element-wise operators on V100 and use the methodology presented in the paper to predict the latency for other GPUs. For V100, we use the latency measured on P100.
For \micro, following the method described in the paper, we apply linear regression between GPU memory bandwidth and achieved FLOPS performance. We then infer the achieved FLOPS performance of the unseen GPU using its memory bandwidth.

\niparagraph{Training the \neusight~predictor.}
We train the MLPs using the AdamW optimizer with L2 regularization for 100 epochs, with batch sizes of ranging from 16 to 128.
For Habitat predictors, the learning rate is set to $5e^{-5}$ for highest accuracy. For \neusight, we use a range of learning rates from $1e^{-6}$ to $5e^{-3}$, tailored to different predictors within the model.
We use mean absolute percentage error for Habitat and symmetric mean absolute percentage error \cite{smape} for \neusight~as the loss functions.

\niparagraph{Tile size.}
We utilize the PyTorch Profiler \cite{torchprof} to obtain the kernel names and the number of thread blocks. For matrix multiplications, we deduce tile sizes from the metadata appended to the kernel names. For other kernels, we use the number of thread blocks to infer tile sizes.
We record the kernel name, input dimensions, GPU features, and tile sizes in a database during training. For prediction, we estimate tile sizes by finding the closest match in the database based on the kernel name, input dimensions, and GPU features.

\niparagraph{DNN workloads evaluated.}
Table \ref{tab:workloads_apdx} shows the 6 commonly used models released between 2018 and 2022. 
For the Switch Transformer, which is a mixture of experts model, a 4-expert configuration is employed.
For each of these models, we measure the inference latency for a binary classification task using BERT, and for a text generation task using GPT-2, GPT-3, OPT, and Switch Transformer. 
We use the time to generate the first token as the latency metric for text generation tasks. We also report and compare the per-iteration training time, including a single forward and backward pass. 
Predictions for each model are then compared with the measured latencies to evaluate our framework.
Due to the high memory requirement of training, we measure training latency on GPUs with at least 24~GB of HBM.

\subsection{Single-Device Execution Results}

\niparagraph{End-to-end mean percentage error.}
Figure \ref{fig:main_eval} shows the inference and training latency prediction error of \neusight~and the baselines on 6 deep learning models with varying batch sizes executed on 8 GPUs in percentage error.
Instances that include out-of-distribution GPUs or models are highlighted. For example, H100 is an out-of-distribution GPU as is not included in the training set. 
Similarly, GPT3-2.7B is an out-of-distribution model as it includes BMM with one of the operand dimensions equal to 2048, which is larger than the 1024 used in the training set.

Across all GPUs and models, \neusight~shows a modest average percentage error compared to all other baselines. Specifically, the roofline analysis shows an average error of 31.2\% and 31.9\%, Habitat shows 220.9\% and 725.8\%, and \micro~shows 61.2\% and 58.3\%, for inference and training, respectively. 
In contrast, \neusight~shows a percentage error of only 9.7\% for inference and 7.3\% for training workloads.
\revision{Smaller models relative to GPU capacity, like BERT on H100, tend to show higher error rates due to library overheads. Nonetheless, NeuSight’s kernel tiling approach still achieves reasonable accuracy.}

\revision{
The strengths of \neusight~are highlighted on out-of-distribution GPUs, where it achieves an average latency prediction error of 8.1\% and a maximum error of 28.2\%.
In comparison, Habitat and \micro~show average errors of 724.3\% and 94.0\%, with maximum errors of 4529.9\% and 435.9\%, respectively, when trained on the same dataset with \neusight.
}
%
The main reason for the lower absolute percentage error of \neusight~is that it does not directly use the MLP to make latency predictions, instead applies GPU architecture and execution insights atop the machine learning predictions. 
In the case of newer generations GPUs like H100, which exhibit features outside the distribution seen in the training set, Habitat's and \micro~observe higher errors.
The lower error rate of \neusight~showcases its robustness in handling out-of-distribution features and making accurate predictions even on unseen GPUs.

\begin{figure}[t]
\centering
\includegraphics[width=0.9\columnwidth]{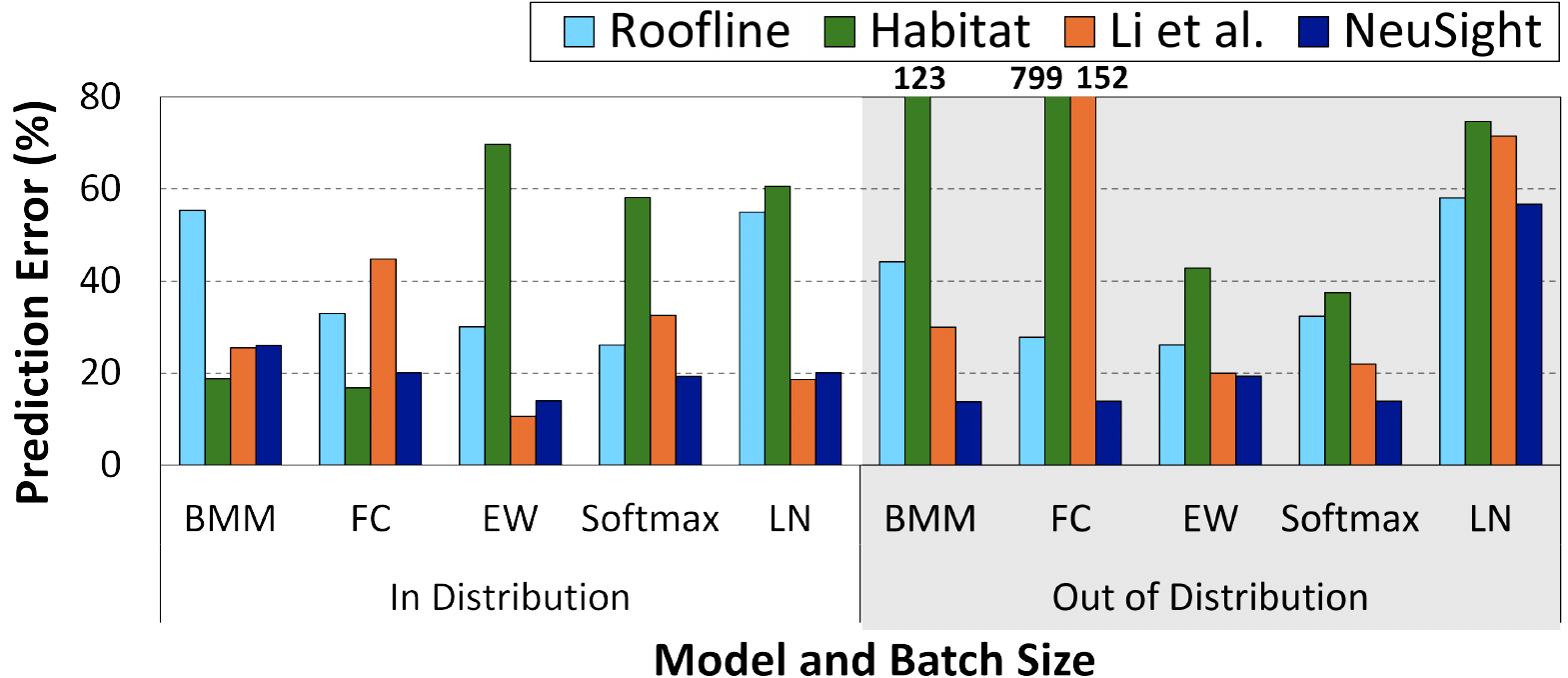}
\vspace{-1ex}
\caption{Prediction percentage error for each operator type (BMM, Fully-Connected (FC), Element-Wise (EW), Softmax, Layer Normalization (LN)), averaged over all the evaluated workloads.}
\vspace{-2ex}
\label{fig:perop}
\end{figure}

\niparagraph{Per-operator prediction error.}
Figure \ref{fig:perop} shows the prediction error for each kernel type (BMM, fully-connected, element-wise, softmax, layer normalization), averaged across all workloads evaluated in Figure \ref{fig:main_eval}.
The roofline analysis consistently shows a higher percentage error compared to \neusight~for both in-distribution and out-of-distribution data points, with an average percentage error of 34\%. 
The accuracy of Habitat and \micro~degrades on out-of-distribution data points, particularly for BMM and fully-connected operators.
Habitat and \micro~show percentage error of 123.2\% and 30.0\% for BMM, and 799.3\% and 152.6\% for fully-connected operators, respectively. 
In contrast, \neusight~shows better accuracy than roofline analysis, Habitat and \micro, with a percentage error of 13.8\% for BMM operators and 13.9\% for fully-connected operators.
%
%

\revision{
Table \ref{tab:contribution} shows the contribution of each layer type to the end-to-end latency.
Since GEMM operations contribute the most to the overall latency, accurate GEMM predictions of \neusight~are crucial for maintaining low error rates in end-to-end performance predictions.
We observed higher error rates for layer normalization due to its generally short latency, which makes accurate predictions challenging. 
However, as layer normalization contributes only up-to 2\% to the overall latency, its higher error does not significantly affect \neusight's end-to-end prediction accuracy.
}

\begin{footnotesize}
\newcommand\ExtraSep
{\dimexpr\cmidrulewidth+\aboverulesep+\belowrulesep\relax}

\newcolumntype{?}{!{\vrule width 2pt}}
\setlength\extrarowheight{3pt}

\begin{table}
\centering
\caption{Per-operator contribution to the overall latency for various models on H100 for model inference. The contributions from different types of operators are shown as percentages.}
\revisionfig{
\resizebox{1.0\columnwidth}{!}
{
\begin{tabular}{llllllll}
\hline
\textbf{Model}           & \textbf{Batch Size} & \textbf{BMM} & \textbf{LINEAR} & \textbf{EW} & \textbf{SOFTMAX} & \textbf{LN} & \textbf{OTHERS} \\ \hline
\textbf{BERT-Large} & 16                  & 12\%         & 74\%            & 10\%        & 3\%              & 2\%         & 0\%             \\ \hline
\textbf{GPT2-Large} & 4                   & 11.8\%       & 62.8\%          & 14.9\%      & 4.0\%            & 1.1\%       & 5.3\%           \\ \hline
\textbf{OPT13}      & 2                   & 13.1\%       & 62.9\%          & 9.6\%       & 5.9\%            & 0.6\%       & 7.8\%           \\ \hline
\textbf{GPT3-XL}    & 2                   & 9.5\%        & 76.4\%          & 7.9\%       & 2.5\%            & 0.4\%       & 3.3\%           \\ \hline
\end{tabular}
}
}

\label{tab:contribution}

\end{table}
\end{footnotesize}

\begin{figure}[t!]
\centering
\vspace{-2ex}
\subfloat[\label{fig:amdinf} Inference latency prediction error]{\revisionfig{\includegraphics[width=0.9\columnwidth]{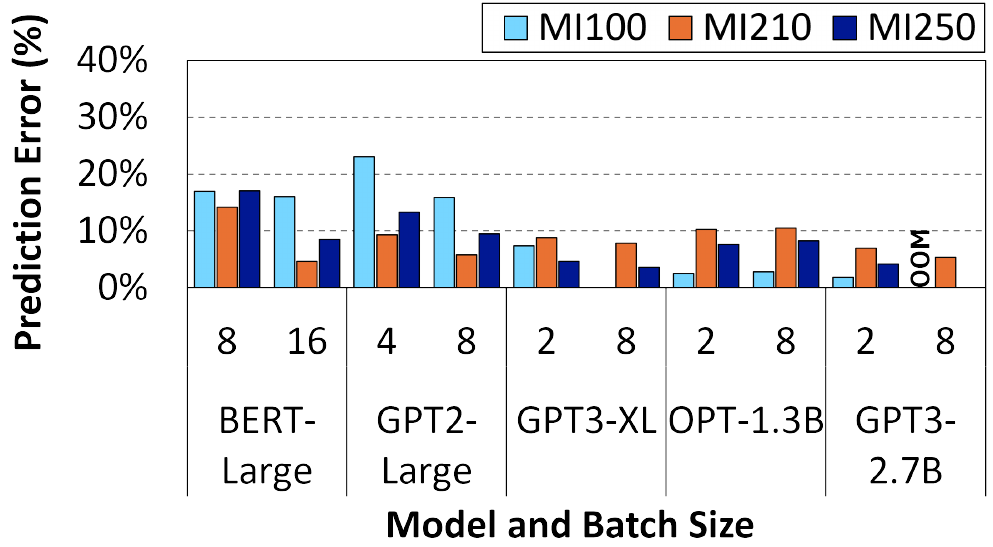}}}
\\
\subfloat[\label{fig:amdtrain}Training latency prediction error]{\revisionfig{\includegraphics[width=0.9\columnwidth]{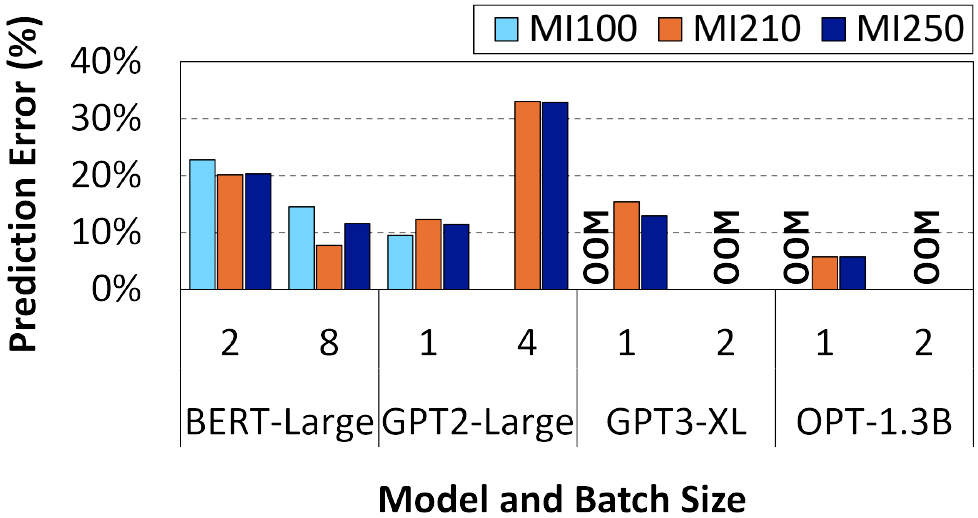}}}
\vspace{-2ex}
\caption{The inference and training latency prediction percentage error of \neusight~for various models across different batch sizes on AMD MI100 and MI210 GPUs.}
\vspace{-2ex}
\label{fig:amdgpueval}
\end{figure}

\niparagraph{GPU across vendors (AMD).}
\revision{Modern GPUs, including those from AMD, share many execution and deep learning concepts discussed in this paper. For instance, AMD's CDNA architecture uses multiple identical compute units as its core building blocks, similar to NVIDIA's Streaming Multiprocessors (SMs), and AMD's HIP programming model employs workgroups similar to CUDA's thread blocks\cite{amd_gpus}.
We trained \neusight~using data from MI100 and MI210 and evaluated its prediction accuracy on the MI250.
We evaluated the prediction accuracy of \neusight~across 5 model architectures varying batch sizes, including GPT3-XL, GPT3-2.7B, and OPT-1.3B which contain out-of-distribution operations.
Figure \ref{fig:amdgpueval} presents the AMD GPU evaluation results.
Across all tested models, \neusight~achieves an average percentage error of 8.8\% for inference and 15.7\% for training, demonstrating its ability to generalize across different GPU vendors.}

\begin{scriptsize}
\newcommand\ExtraSep
{\dimexpr\cmidrulewidth+\aboverulesep+\belowrulesep\relax}
\newcolumntype{?}{!{\vrule width 2pt}}
\setlength\extrarowheight{3pt}

\begin{table*}[t]

\caption{The inference latency prediction \textcolor{brickred}{percentage error} of \neusight~with operator fusion with varying batch sizes.}

\centering

\begin{tabular}{lllllllll}
\hline
           &  \textbf{Batch} &                       & \textbf{L4}              &                & \textbf{A100-40GB}      &                 & \textbf{H100}           &                \\ \cline{4-9} 
           &  \textbf{Size}  &                            & Non-fused       & Fused          & Non-fused      & Fused           & Non-fused      & Fused          \\ \hline
\textbf{BERT-Large} & 8          & Measured Latency (ms)      & 409.9           & 344.1          & 205.5          & 174.8          & 69.8           & 65.0           \\ \cline{3-9} 
           &          & NeuSight Prediction   (ms) & 350.1 \textcolor{brickred}{(14.6\%)}  & 294.3 \textcolor{brickred}{(14.5\%)} & 216.7 \textcolor{brickred}{(5.4\%)}  & 204.8 \textcolor{brickred}{(17.2\%)}   & 86.2 \textcolor{brickred}{(23.4\%)}  & 80.9 \textcolor{brickred}{(24.6\%)}  \\ \cline{2-9} 
           & 16         & Measured Latency (ms)      & 855.4           & 724.7          & 396.5          & 345.2           & 136.8          & 127.0          \\ \cline{3-9} 
           &            & NeuSight Prediction   (ms) & 681.1 \textcolor{brickred}{(20.4\%)}  & 573.4 \textcolor{brickred}{(20.9\%)} & 414.1 \textcolor{brickred}{(4.5\%)}  & 391.5 \textcolor{brickred}{(13.4\%)}   & 154.8 \textcolor{brickred}{(13.2\%)} & 144.7 \textcolor{brickred}{(14.0\%)} \\ \hline
\textbf{GPT2-Large} & 4          & Measured Latency (ms)      & 1276.3          & 886.8          & 535.8          & 477.6           & 215.0          & 174.7          \\ \cline{3-9} 
           &            & NeuSight Prediction   (ms) & 1135.9 \textcolor{brickred}{(11.0\%)} & 952.5 \textcolor{brickred}{(7.4\%)}  & 576.2 \textcolor{brickred}{(7.5\%)}  & 537.6 \textcolor{brickred}{(12.6\%)}   & 208.4 \textcolor{brickred}{(3.1\%)}  & 190.6 \textcolor{brickred}{(9.1\%)}  \\ \cline{2-9} 
           & 8          & Measured Latency (ms)      & 2563.0          & 1775.1         & 1084.7         & 948.5           & 414.3          & 343.4          \\ \cline{3-9} 
           &            & NeuSight Prediction   (ms) & 2299.0 \textcolor{brickred}{(10.3\%)}  & 1943.8 \textcolor{brickred}{(9.5\%)} & 1179.4 \textcolor{brickred}{(8.7\%)} & 1107.9 \textcolor{brickred}{(16.8\%)} & 442.5 \textcolor{brickred}{(6.8\%)} & 409.9 \textcolor{brickred}{(19.4\%)} \\ \hline
\end{tabular}
\label{tab:fusion}
\end{table*}

\end{scriptsize}

\niparagraph{Operator fusion.}
Table \ref{tab:fusion} shows the inference latency prediction error of BERT-Large and GPT2-Large with operator fusion on L4, A100 (40GB), and H100. Using torch.compile \cite{torchcompile}, the built-in operator fusion tool in PyTorch, we fused the supported operators and measured the latency of the fused models. 
\neusight~predicts latency with a 15.7\% error across all fused models, 18.9\% for BERT-Large, and 12.5\% for GPT2-Large.
This implies the generalizability of \neusight~to capture the common optimizations employed for deep learning kernels, as it distills the execution to tile and waves.

\begin{figure}[t]
\centering
\revisionfig{\includegraphics[width=0.9\columnwidth]{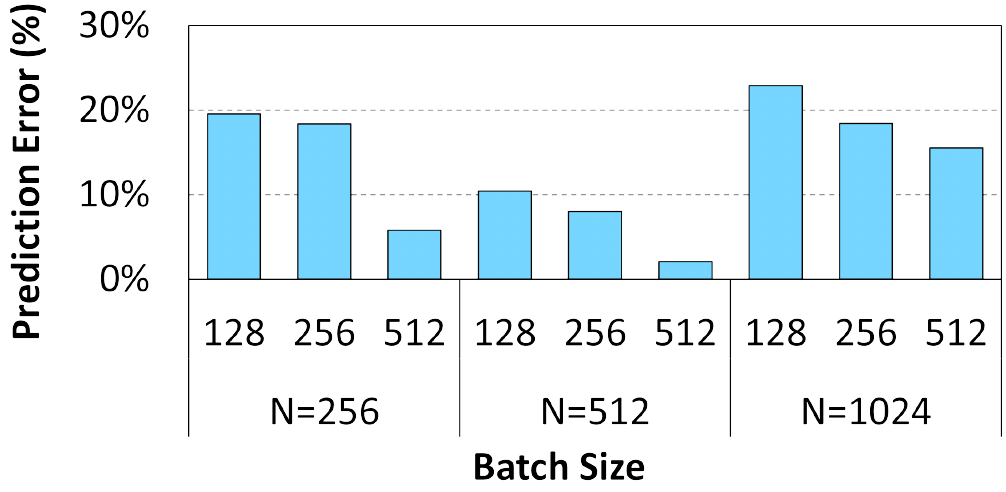}}
\vspace{-1ex}
\caption{Performance of (NxN) x (NxN) matrix multiplication on H100 GPUs using FP16 numeric type and Tensor Cores across different batch sizes.}
\vspace{-2ex}
\label{fig:tensorcore}
\end{figure}

\niparagraph{New numerical type and hardware unit.}
\revision{Even in the case of the introduction of new hardware units and software changes, \neusight~can be adapted to predict the performance of unseen hardware units.
This is because core concepts such as multiple identical processing units, tiled execution, and fundamental performance equations remain relevant.
We demonstrate this adaptability in this experiment, where \neusight~is tested on a new numerical type (FP16) and a new hardware unit (Tensor Core).
Figure \ref{fig:tensorcore} presents the latency predictions by \neusight~for FP16 matrix-multiplication on H100 Tensor Cores, illustrating the versatility of \neusight~on new hardware.
To adapt to changes in hardware and software features, we adjust the input features, accounting for the reduced memory requirement due to lower precision and the higher peak FLOPS to reflect the enhanced compute capacity of the Tensor Core.
\neusight~predicts the latency of the batched matrix multiplication with an average error rate of 13\%, showcasing its reliability even with these hardware and software changes.}

\begin{scriptsize}
\newcommand\ExtraSep
{\dimexpr\cmidrulewidth+\aboverulesep+\belowrulesep\relax}
\newcolumntype{?}{!{\vrule width 2pt}}
\setlength\extrarowheight{3pt}

\begin{table*}[t]

\centering
\caption{The distributed training latency prediction \textcolor{brickred}{percentage error} of \neusight~on H100 and A100 GPU servers training GPT2-Large and GPT3-XL with different global batch sizes. Models resulting in out-of-memory errors are marked as OOM.}
\begin{tabular}{lllllllll}
\hline
           & \textbf{Global} & & \multicolumn{3}{l}{\textbf{A100-40GB x 4 (NVLink)}} & \multicolumn{3}{l}{\textbf{H100 x 4 (DGX Box)}}   \\ \cline{4-9}
           & \textbf{Batch Size} &              & Data Parallel & Tensor Parallel & Pipeline Parallel & Data Parallel & Tensor Parallel & Pipeline Parallel \\ \hline
\textbf{GPT2-Large} & 4          & Measured Latency (ms)    & 452.3         & 484.9           & 1597.6             & 194.9         & 211.1            & 637.5             \\ \cline{3-9}
           &            & NeuSight Prediction (ms) & 510.5 \textcolor{brickred}{(12.9\%)} & 525.6 \textcolor{brickred}{(8.4\%)}   & 1761.6 \textcolor{brickred}{(10.3\%)}     & 214.1 \textcolor{brickred}{(9.9\%)} & 228.1 \textcolor{brickred}{(8.0\%)}    & 650.2 \textcolor{brickred}{(2.0\%)}     \\ \cline{2-9} 
           & 16         & Measured Latency (ms)    & OOM           & OOM             & OOM               & 642.1         & 729.7              & 2384.0            \\  \cline{3-9}
           &            & NeuSight Prediction (ms) &               &                 &                   & 649.6 \textcolor{brickred}{(1.2\%)} & 825.1 \textcolor{brickred}{(13.1\%)}   & 2638.4 \textcolor{brickred}{(10.7\%)}   \\ \hline 
\textbf{GPT3-XL}    & 4          & Measured Latency (ms)    & OOM           & OOM             & OOM               & OOM           & 1010.5            & 3554.4            \\  \cline{3-9}
           &            & NeuSight Prediction (ms) &               &                 &                   &               & 1047.6 \textcolor{brickred}{(3.7\%)}   & 3717.0 \textcolor{brickred}{(4.6\%)}   \\ \hline
\end{tabular}
\label{tab:distributed}

\end{table*}

\end{scriptsize}

\subsection{Distributed Execution Results}

\niparagraph{Single-server execution.} Table \ref{tab:distributed} shows the training latency predictions for distributed execution on two types of GPU servers: one with 4 A100s-40GB connected via NVLink and 4 H100s in a DGX Box~\cite{h100dgx}, training GPT2-Large and GPT3-XL.
The A100 GPUs are connected in a mesh topology with each GPU using 12 NVLinks, providing a bi-directional bandwidth of 600GB/s. In the H100 DGX system, each GPU is connected with 18 NVLinks, providing a bi-directional bandwidth of 900GB/s. Both configurations enable full-bandwidth communication between any two GPUs.
We evaluate \neusight~on data, tensor model, and pipeline parallel execution, albeit individually.
This is because \neusight~aims to also estimate the latency for a wide range of models, including the ones which cannot fit on a single device due to memory, thus require tensor model or pipeline to split the DNN graph.
For each method, we split the execution across the 4 GPUs, and use a single micro-batch.

Across all the models, \neusight~predicts the latency of distributed training with an average error of 7.7\%, including 6.7\% for the H100 server and 10.5\% for the A100 server. 
This demonstrates that \neusight~predicts effectively for per-kernel, per-GPU, and distributed execution on various GPUs and server types.
This is mainly due to the robust underlying mechanism of kernel estimation, that can be extended with the network level estimates to forecast distributed execution latencies.

\begin{scriptsize}
\newcommand\ExtraSep
{\dimexpr\cmidrulewidth+\aboverulesep+\belowrulesep\relax}
\newcolumntype{?}{!{\vrule width 2pt}}
\setlength\extrarowheight{3pt}

\begin{table}[t]

\centering
\caption{\neusight~prediction for multi-node distributed execution of GPT-3 on H100 GPUs.}
\revisionfig{
\resizebox{1.0\columnwidth}{!}
{
\begin{tabular}{llllll}
\hline
\textbf{\# Nodes}                 & \textbf{1} & \textbf{4} & \textbf{384} & \textbf{768} & 3840    \\ \hline
\textbf{NeuSight Prediction (ms)} & 1514.9     & 1836.7     & 12028.3      & 12135.5      & 12564.6 \\ \hline
\end{tabular}
\label{tab:multinode}
}
}

\end{table}

\end{scriptsize}

\niparagraph{Multi-node distributed execution.}
\revision{
Extending beyond single-server execution, \neusight~can be used with networking simulators to predict multi-node execution where GPUs communicate across different servers. 
While simulating multi-server networks is well handled by tools such as ASTRA-Sim~\cite{astrasim} and ns-3~\cite{ns3}, \neusight's predictions can be integrated with these simulators to forecast multi-node distributed training.
To demonstrate this capability, we provide estimation results for executing GPT-3 across 1 to 3840 nodes for a single iteration.
The experimental infrastructure consists of nodes equipped with 8 H100 GPUs each, interconnected via a multi-level fat-tree network using InfiniBand switches, with a link bandwidth of 100 Gbps.
Nodes are connected hierarchically across levels consisting of 4, 384, 768, and up to 3840 nodes.
Each DGX server provides GPU-to-GPU bandwidth of 900GB/s~\cite{h100dgx}. 
For the distributed training configuration, we use an 8-wide Tensor Model parallel setup within each server, while we use data parallelism across the nodes.
The per-node batch size is set to 8, and the global batch size scales as 8 times the number of nodes.
\neusight~handles per-device predictions, while the analytical network performance model from the previous subsection estimates the communication latency for AllReduce operations across the switched network. 
Due to resource constraints, we cannot validate this prediction on a real cluster of this size.}

\section{Conclusion}
In this work, we present a novel framework, \neusight, that forecasts the performance of deep learning execution on GPU servers.
In developing \neusight, we leverage the execution deep learning kernel on a GPU architecture such as tiled execution and use this insight to decompose the prediction problem to smaller sub-problems, i.e., per-tile prediction. 
For each tile, we use machine learning models to capture the complex non-linear relationship between achieved throughput and input features, and impose performance laws to bound the predicted latency.
Results on a variety of models and GPUs, show that \neusight~can provide better predictions than prior work in the area. We open-sourced \neusight~to benefit broader research community. \footnote{\neusight~is available at \href{https://github.com/sitar-lab/NeuSight}{github.com/sitar-lab/NeuSight}}
\section{Acknowledgements}

We thank the anonymous reviewers for their insightful comments. 
This research was supported through computational resources provided by Partnership for an Advanced Computing Environment (PACE) at Georgia Tech, Institute for Data Engineering and Science (IDEaS) at Georgia Tech, Google Cloud, and AMD HPC Fund Research Cloud.
This work is partially supported by Gifts from Google and AMD. The views and conclusions contained herein are those of the authors. They should not be interpreted as representing the official policies or endorsements, either expressed or implied, of Georgia Tech or Meta.

\appendix
\section{Artifact Appendix}

\subsection{Abstract}

The artifact includes source code, Python scripts, and prediction datasets required to reproduce the error rates of different ML models on various GPUs, as described in the Evaluation section (Section 6).

\subsection{Artifact check-list (meta-information)}

{\small
\begin{itemize}
  \item {\bf Algorithm:} ML-based GPU execution time predictor
  
  \item {\bf Program:} Python, PyTorch
  
  \item {\bf Model:} Transformer models from HuggingFace
  
  \item {\bf Data set:} Measured latencies of single kernels and end-to-end ML model executions collected from 8 NVIDIA GPUs and 3 AMD GPUs (available in the GitHub repository)
  
  \item {\bf Run-time environment:}
  \begin{itemize}
    \item Ubuntu 20.04, Python 3.9, PyTorch 2.1.0 or similar 
    \item (If collecting datasets from scratch) PyTorch 2.1 + CUDA 12.1 and PyTorch 2.4 + RoCM 6.1
  \end{itemize}
  
  \item {\bf Hardware:} 
  \begin{itemize}
    \item Any GPU supporting PyTorch 2.1 or similar 
    \item (If collecting datasets from scratch) P4, P100, V100, T4, A100, L4, H100, MI100, MI200, and MI250
  \end{itemize}
  
  \item {\bf Execution:} (If collecting datasets from scratch) Sole GPU usage recommended for profiling
  
  \item {\bf Metrics:} Predicted latency and prediction error
  
  \item {\bf Output:} CSV-formatted predicted latency and prediction error values used for tables and figures in the Evaluation section (Section 6)
  
  \item {\bf Disk space required:} \textasciitilde50 GB
  
  \item {\bf Time to prepare workflow:} \textasciitilde1 hour
  
  \item {\bf Time to complete experiments:} \textasciitilde1 hour using the provided dataset; \textasciitilde10 hours to collect datasets, labels, and traces from scratch
  
  \item {\bf Publicly available:} Yes
\end{itemize}
}

\subsection{Description}

\subsubsection{How to access} 
The scripts and code are available at \url{https://github.com/sitar-lab/NeuSight}. Documentation and setup instructions for artifact evaluation are included in \texttt{scripts/asplos/README}.

\subsubsection{Hardware dependencies}
Experiments were conducted on P4, P100, V100, T4, A100, L4, H100, MI100, MI200, and MI250.  
If using the provided dataset, any GPU supporting PyTorch 2.1 or similar is sufficient.

\subsubsection{Software dependencies}
\begin{itemize}
  \item {\bf For NVIDIA GPUs:} Ubuntu 20.04 or similar, Python 3.9, PyTorch 2.1.0, CUDA Toolkit 12.1
  \item {\bf For AMD GPUs:} RHEL 9.4 or similar, Python 3.12.5, PyTorch 2.4.1, RoCM 6.1
\end{itemize}

\subsubsection{Data sets}
The publicly available dataset included in the GitHub repository contains operator latencies and ground truth latency measurements of various ML models across different GPUs.

\subsection{Installation and Testing}

\subsubsection{Installation}
\begin{verbatim}
git clone https://github.com/sitar-lab/NeuSight
cd NeuSight
pip install -e .
\end{verbatim}

\subsubsection{Basic Test}
\begin{verbatim}
bash scripts/example/gpt3_inference_h100.sh
\end{verbatim}
The printed output should be approximately \textasciitilde670 ms.

\subsection{Experiment workflow}

1. (Optional) Run scripts to collect datasets from scratch. \\
2. (Optional) Train NeuSight with the collected dataset. \\
3. Use the provided evaluation scripts to predict model execution latencies on GPUs evaluated in the paper. \\
4. Compare predictions against ground truth using the provided \texttt{summary.py} and \texttt{table.py} scripts.

\subsection{Evaluation and expected results}

The artifact reproduces the results shown in Figures 7, 8, and 9, and Tables 7 and 8 of the paper. Data for these figures and tables is available in \texttt{/scripts/asplos/summary}. Note that latency predictions may vary slightly (\textasciitilde10\%) due to the non-deterministic behavior of DNN models.

\bibliographystyle{ACM-Reference-Format}
\balance
\bibliography{references}

\end{document}